\documentclass[10pt,twocolumn,letterpaper]{article}

\usepackage{iccv}
\usepackage{times}
\usepackage{epsfig}
\usepackage{graphicx}
\usepackage{amsmath}
\usepackage{amssymb}
\usepackage[pdftex]{xcolor}

\usepackage{framed,multirow}
\usepackage{amsthm}
\usepackage{comment}
\usepackage{booktabs}
\theoremstyle{plain}
\newtheorem{pro}{Proposition}

\usepackage[pagebackref=true,breaklinks=true,letterpaper=true,colorlinks,bookmarks=false]{hyperref}

\iccvfinalcopy 

\def\iccvPaperID{1548} 

\ificcvfinal\fi

\begin{document}

\title{NLOS-NeuS: Non-line-of-sight Neural Implicit Surface}

\author{Yuki Fujimura$^1$ 
\and 
Takahiro Kushida$^1$ 
\and 
Takuya Funatomi$^1$
\and 
Yasuhiro Mukaigawa$^1$ \\
$^1$Nara Institute of Science and Technology, Japan\\
{\tt\small \{fujimura.yuki,kushida.takahiro.kh3,funatomi,mukaigawa\}@is.naist.jp}
}

\maketitle
\ificcvfinal\thispagestyle{empty}\fi

\begin{abstract}
Non-line-of-sight (NLOS) imaging is conducted to infer invisible scenes from indirect light on visible objects.
The neural transient field (NeTF) was proposed for representing scenes as neural radiance fields in NLOS scenes.
We propose NLOS neural implicit surface (NLOS-NeuS), which extends the NeTF to neural implicit surfaces with a signed distance function (SDF) for reconstructing three-dimensional surfaces in NLOS scenes.
We introduce two constraints as loss functions for correctly learning an SDF to avoid non-zero level-set surfaces.
We also introduce a lower bound constraint of an SDF based on the geometry of the first-returning photons.
The experimental results indicate that these constraints are essential for learning a correct SDF in NLOS scenes.
Compared with previous methods with discretized representation, NLOS-NeuS with the neural continuous representation enables us to reconstruct smooth surfaces while preserving fine details in NLOS scenes.
To the best of our knowledge, this is the first study on neural implicit surfaces with volume rendering in NLOS scenes.
\end{abstract}


\begin{figure}[tb]
\centering
\includegraphics[width=0.45\textwidth]{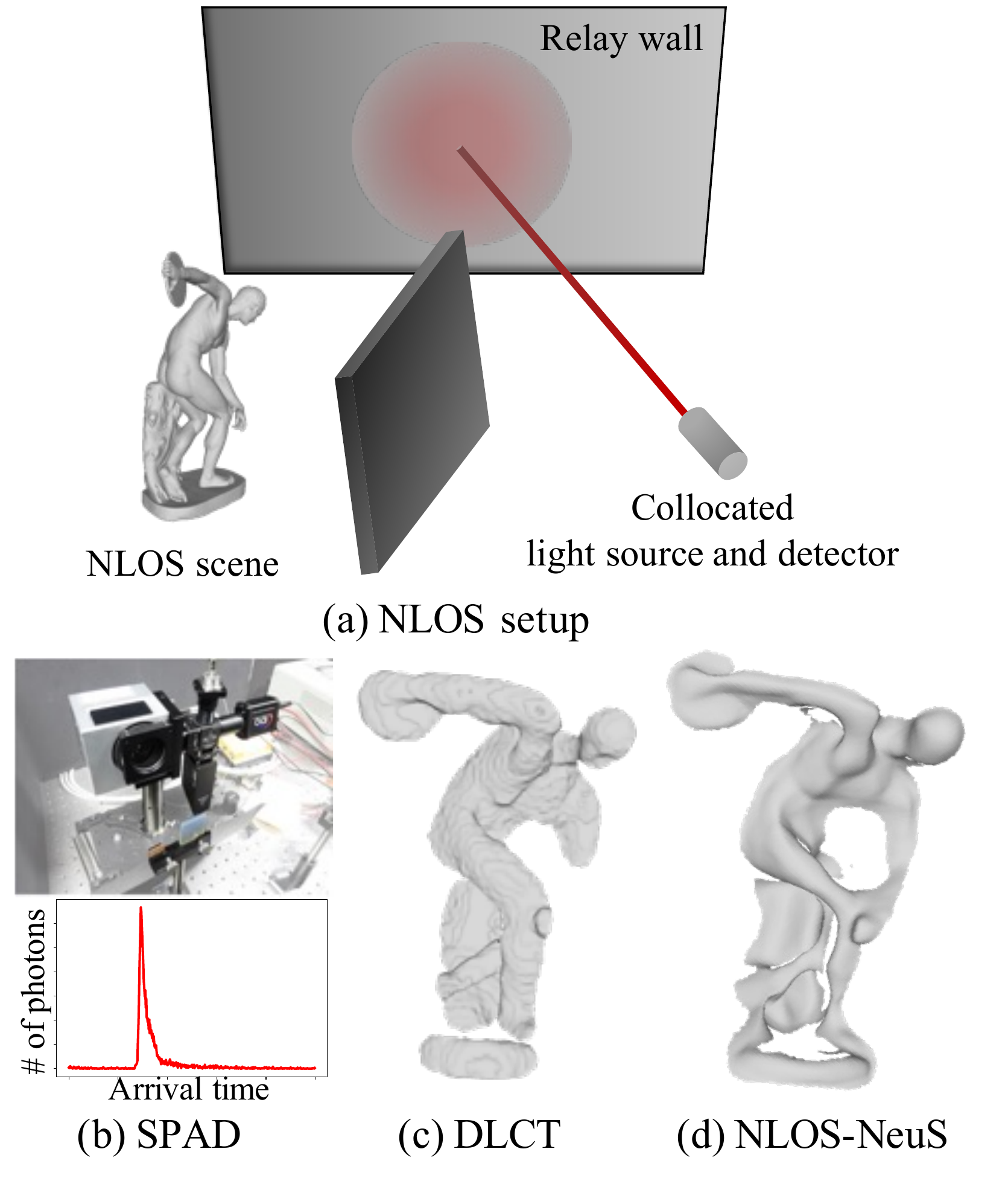}
\caption{(a) Confocal NLOS setup, (b) SPAD system and measured transients, (c) NLOS surface reconstruction with DLCT \cite{Young2020} and (d) with our NLOS-NeuS. Neural implicit surface representation can enhance quality of reconstructed surfaces in NLOS scenes.}
\label{fig:nlos_setup}
\end{figure}

\section{Introduction} \label{sec:introduction}
Many computer-vision applications are being integrated into society.
However, most are intended for visible scenes from cameras.
Non-line-of-sight (NLOS) imaging \cite{Kirmani2011,Velten2012} is used to infer invisible scenes occluded from the cameras.
Figure \ref{fig:nlos_setup}(a) shows a typical confocal NLOS setup \cite{OToole2018}, in which an ultra-fast light source and time-resolved detector are collocated and the detector can only see a diffuse wall.
Pulsed light from the light source is emitted into the visible wall, and reflected light on the wall reaches an NLOS scene.
Reflected light in the NLOS scene then bounces back to the detector via the relay wall.
A time-resolved detectors for such NLOS imaging is a single photon avalanche diode (SPAD) \cite{Buttafava2015} (top in Fig \ref{fig:nlos_setup}(b)).
A SPAD is capable of time-correlated single photon counting with pico-second resolution \cite{Warburton2016}.
Measured data with such a time-resolved detector are called transients, in which the number of counted photons or intensity is recorded at each arrival time (bottom in Fig. \ref{fig:nlos_setup}(b)).
One approach for NLOS imaging is to use the transients on the relay wall as input.
We aim to reconstruct three-dimensional (3D) surfaces in NLOS scenes from the transients, similar to previous studies \cite{Iseringhausen2020,Plack2023,Tsai2017,Tsai2019,Xin2019,Young2020}.

We extend the neural transient field (NeTF) \cite{Shen2021} for neural implicit surface representation in NLOS scenes.
The neural radiance field (NeRF) \cite{Mildenhall2020} is a powerful paradigm for scene representation, in which density and view-dependent color at each scene position are continuously parameterized with a multi-layer perceptron (MLP), which is optimized by reconstructing input multi-view images with volume rendering.
The NeTF is the extended version of the NeRF for transient measurement in an NLOS setup.
However, the NeTF (and also the NeRF) does not support geometric representation, which is necessary for accurate 3D reconstruction.
We incorporate the framework of a neural implicit surface with volume rendering such as NeuS \cite{Wang2021neus} and VolSDF \cite{Yariv2021} into the NeTF for NLOS neural implicit surface reconstruction, i.e., {\it NLOS-NeuS}.

We argue that a simple extension of the NeTF to neural implicit surface representation with a signed distance function (SDF) causes a non-zero level-set surface due to an under-constrained NLOS setup.
We introduce effective training losses to avoid a non-zero level-set surface.
Figure \ref{fig:nlos_setup}(d) shows an example of the surface reconstruction of an occluded object with NLOS-NeuS.
Compared with the results of the state-of-the art NLOS surface-reconstruction method \cite{Young2020} (Fig. \ref{fig:nlos_setup}(c)), the fine details of the target object can be reconstructed because of the nature of its continuous representation.
To the best of our knowledge, this is the first study on a neural implicit surface with volume rendering in NLOS scenes. 
\section{Related work}
\paragraph{Non-line-of-sight imaging}
NLOS imaging is attracting much more attention with the development of computational imaging devices \cite{Kirmani2011,Velten2012}.
While the typical inputs of NLOS-imaging methods are transients on a diffuse relay wall, output scene representations differ depending on applications and methodology.

A voxel grid is one of the most commonly used representations \cite{Arellano2017,Lindell2019,Liu2019,OToole2018,Velten2012}.
For example, with back-projection-based methods \cite{Arellano2017,LaManna2019,Velten2012}, the measured intensity is back-projected to each voxel to estimate the probabilities of object existence.
O’Toole et al. \cite{OToole2018} proposed the light-corn transform (LCT), with which a closed-form solution is derived under volumetric albedo representation.
Lindell et al. \cite{Lindell2019} formulated the NLOS problem as f-k migration in seismology, where electromagnetic radiation at each scene-grid point is recovered with the Fast Fourier Transform.
However, such discrete representations are limited for representing scenes with fine details due to memory cost.
Shen et al. \cite{Shen2021} proposed the NeTF, with which volumetric density and reflectance are implicitly modeled by a continuous MLP with arbitrary resolution.
Mu et al. \cite{Mu2022} extended the NeTF to feedforward inference, which enables fast NLOS imaging with a non-confocal setup.

Although these methods model object existence or volumetric albedo in NLOS scenes, accurate 3D reconstruction requires explicit geometric representations,
e.g., some methods incorporate surface normals into voxel representations \cite{Young2020}, directly estimate a point cloud \cite{Xin2019}, or optimize object surface with a differentiable renderer \cite{Iseringhausen2020,Plack2023,Tsai2019}.
Our method extends continuous volumetric representation \cite{Shen2021} with an SDF for surface reconstruction with arbitrary resolution.

\paragraph{Neural implicit surface}
Park et al. \cite{Park2019} and Mescheder et al. \cite{Mescheder2019} respectively proposed the DeepSDF and Occupancy Networks, with which a scene is represented using a SDF and occupancy field parameterized with an MLP. 
In contrast to the traditional discretized representations such as voxels, these implicit representations are memory-efficient, and an object surface can be extracted as level-set at any resolution, which enables dense surface reconstruction from a coarse voxel grid or sparse point cloud with feedforward inference \cite{Chibane2020,Peng2020} or test-time optimization \cite{Atzmon2020,Chabra2020,Ma2022,Williams2022}.

Other research directions include neural implicit surfaces from multi-view 2D images, where the implicit function is optimized by minimizing reconstruction loss between the input and rendered images.
The key to optimization is how to connect the implicit function and surface rendering in a differentiable manner \cite{Niemeyer2020,Yariv2020}.
In contrast to surface-rendering-based methods, Yariv et al. \cite{Yariv2021} and Wang et al. \cite{Wang2021neus} respectively proposed VolSDF and NeuS, with which images are rendered with volume rendering similar to the NeRF \cite{Mildenhall2020}.
Differing from the surface-rendering approach, 
images are rendered with multiple points on a ray with $\alpha$-compositions, which enables back-propagation from not only the surface points but also points far from the surface.
NLOS-NeuS uses a neural implicit surface with volume rendering for NLOS surface reconstruction. 
\section{Method}
This section describes the key points for NLOS-NeuS to correctly learn an SDF in an NLOS scene.
We first explain our problem setting and review the NeTF as preliminary for neural scene representation in an NLOS scene.
We then describe the difficulty in learning an SDF in the under-constrained NLOS setup followed by additional constraints and loss functions for learning an SDF.
At the end of this section, we introduce background rendering in NLOS scenes.

\subsection{Problem setting} \label{sec:problem_setting}
Our goal is to learn two MLPs, $d:\mathbb{R}^3 \to \mathbb{R}$ and $\rho:\mathbb{R}^6 \to \mathbb{R}$ from transients.
The input of $d$ is a scene position $\mathbf{p} = (x,y,z) \in \mathbb{R}^3$ in an NLOS scene to estimate a signed distance, and the inputs of $\rho$ are $\mathbf{p}$ and direction $\mathbf{v}$ from $\mathbf{p}$ to the position of a visible relay wall $\mathbf{p}' = (x',y',z') \in \mathbb{R}^3$ to estimate view-dependent reflectance. 
The parameters of these two MLPs are optimized by reconstructing measured transients on the relay wall with volume rendering. 

The transient at each position on the relay wall is represented with a 1D vector $\tau(\mathbf{p}') \in \mathbb{R}^B$, where $B$ is the number of time bins.
For example, in SPAD measurement, $\tau(\mathbf{p}', t)$ is the number of counted photons, the arrival time of which is $t$.

\subsection{Preliminary}
We first review the NeTF \cite{Shen2021} for NLOS scene representation.
The NeTF also learns two MLPs; one is $\sigma: \mathbb{R}^3 \to \mathbb{R}$ taking as input a position $(x,y,z)$ to estimate density, and the other is $\rho$, which is essentially the same with our MLP to estimate view-dependent reflectance\footnote{In fact, $\rho$ is a 5D function in the NeTF, i.e, the inputs are $(x,y,z)$ and $(\theta, \phi)$.}.
Differing from the NeRF \cite{Mildenhall2020}, where 
scene points are sampled on a ray for volume rendering, the NeTF spherically samples points from the relay wall for rendering the transients as shown in Fig. \ref{fig:netf_sampling}:

\begin{figure}[tb]
\centering
\includegraphics[width=0.4\textwidth]{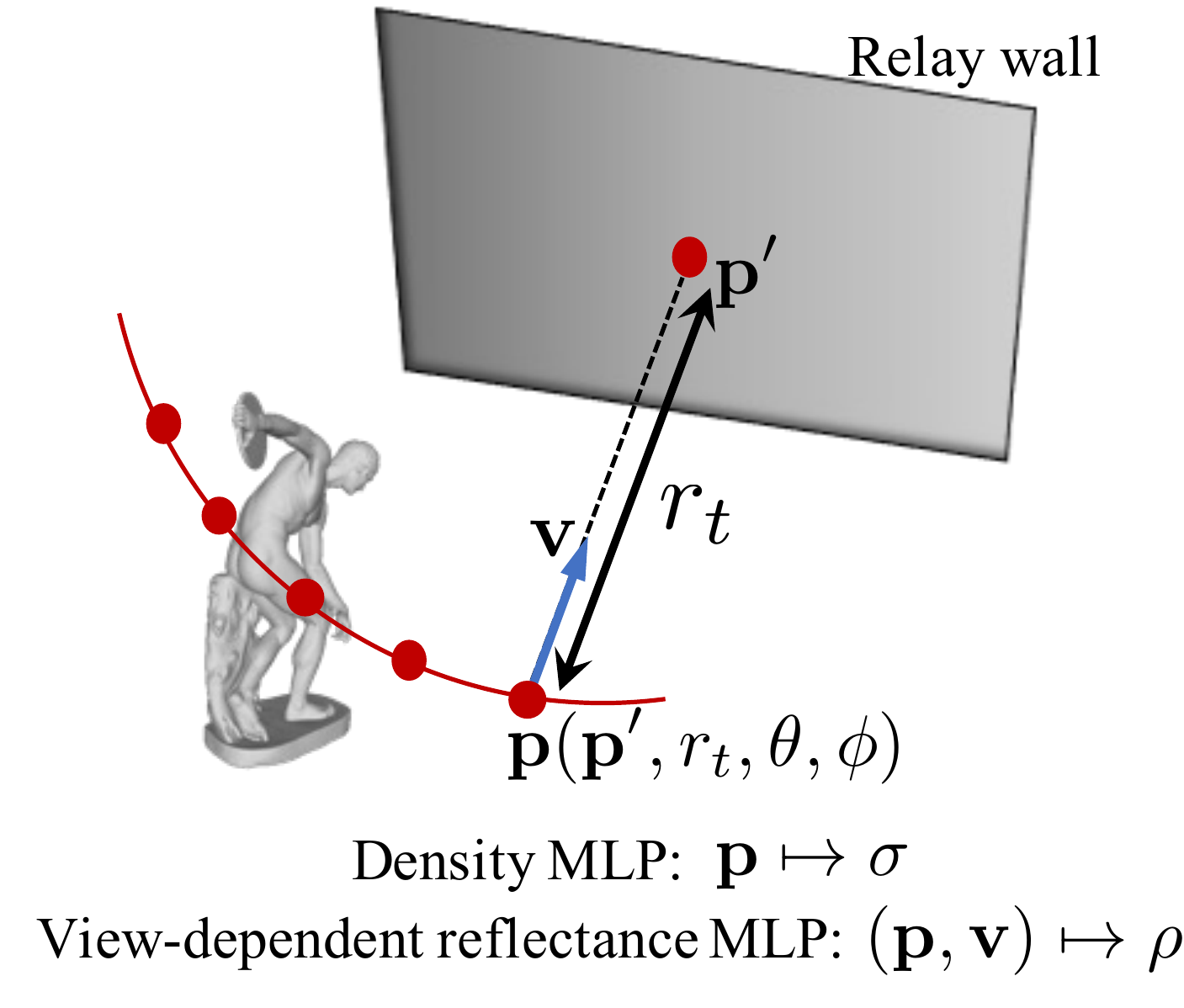}
\caption{Spherical sampling in NeTF \cite{Shen2021} for rendering transients. Scan sphere with radius $r_t$ corresponds to transient at time bin $t$.}
\label{fig:netf_sampling}
\end{figure}

\begin{align}
\mathbf{p}(\mathbf{p}',r_t, \theta, \phi) = \left[
\begin{array}{c}
r_t \sin \theta \cos \phi + x' \\
r_t \sin \theta \sin \phi + y' \\
r_t \cos \theta + z'
\end{array}
\right],
\end{align}
where $r_t$ is a radius corresponding to a transient time bin $t$, and $\theta$ and $\phi$ are elevation and azimuth angles.
A transient $\tau(\mathbf{p}', t)$ is computed as the sum of reflected intensities at sampled points on the scan sphere:
\begin{align}
\tau(\mathbf{p}', t) = \iint A(r_t, \theta) T(\mathbf{p},\mathbf{v})
\sigma (\mathbf{p} ) \rho ( \mathbf{p}, \mathbf{v} ) d\theta d\phi,
\label{eq:rendering_eq}
\end{align}
where $A(r_t,\theta)=\sin \theta / r_t^2$ is an attenuation factor, and $T(\mathbf{p}, \mathbf{v})$\footnote{This transparency is omitted in the implementation of the NeTF.} is the transparency between the object and relay wall.
For simplicity, the arguments of $\mathbf{p}$ and $\mathbf{v}$ are omitted, i.e., $\mathbf{p}(\mathbf{p}',r_t,\theta,\phi)$ and $\mathbf{v}(\mathbf{p}(\mathbf{p}',r_t,\theta,\phi), \mathbf{p}')$.

The parameters of $\sigma$ and $\rho$ are optimized by minimizing the error between measured and rendered transients as follows:
\begin{equation}
\mathcal{L}_\tau = \frac{1}{MB}\sum_{\mathbf{p}',t}(\tau_{m}(\mathbf{p}',t) - \tau(\mathbf{p}',t))^2,
\end{equation}
where $\tau_m$ is the measured transient and $M$ is the number of measured positions on the relay wall.

\subsection{NLOS-NeuS} \label{sec:nlos-surface}
Following previous studies \cite{Wang2021neus,Yariv2021}, NLOS-NeuS uses volume rendering with an SDF.
The MLP $d$ first takes $\mathbf{p}$ as input to estimate a signed distance, and $\rho$ takes $\mathbf{p}$ and $\mathbf{v}$ to estimate
view-dependent reflectance.
The estimated signed distance is then converted to a density for volume rendering.
We use the transformation proposed in StyleSDF \cite{Or-El2022} for computational efficiency with one learnable parameter and without gradient evaluation:
\begin{equation}
\sigma(\mathbf{p}) = \frac{1}{\alpha} Sigmoid(-\frac{d(\mathbf{p})}{\alpha}),
\label{eq:sdf2density}
\end{equation}
where $Sigmoid(\cdot)$ is a sigmoid function and $\alpha > 0$ is a learnable parameter that controls the tightness of the density around the object surface.
After obtaining the density, we can render a transient by using Eq. (\ref{eq:rendering_eq}).
In the implementation, we use the following discrete counterparts \cite{Max1995,Mildenhall2020,Wang2021neus}:
\begin{align}
\tau(\mathbf{p}', t) &= \sum_{\theta,\phi} A(r_t, \theta) w(\mathbf{p},\mathbf{v}) \rho(\mathbf{p},\mathbf{v}) \Delta\theta \Delta \phi, \label{eq:discrete} \\
w(\mathbf{p}, \mathbf{v}) &= \sum_{s=t_{min}}^{t} T_s \Big(1 - \exp\{-\sigma\big(\mathbf{p}(\mathbf{p}',r_s, \theta, \phi)\big)\Delta r_s\}\Big) \label{eq:weight},\\
T_s & = \exp \left\{ -\sum_{u=t_{min}}^{s-1}  \sigma\big(\mathbf{p}(\mathbf{p}',r_u, \theta, \phi)\big) \Delta r_u \right\},
\end{align}
where $w(\mathbf{p}, \mathbf{v})$ corresponds to the blending weights of volumetric albedos on the ray from the relay wall. 

\subsection{Learning zero level-set surface in NLOS setup}\label{sec:nonzero_levelset}
\begin{figure}[tb]
\centering
\includegraphics[width=0.45\textwidth]{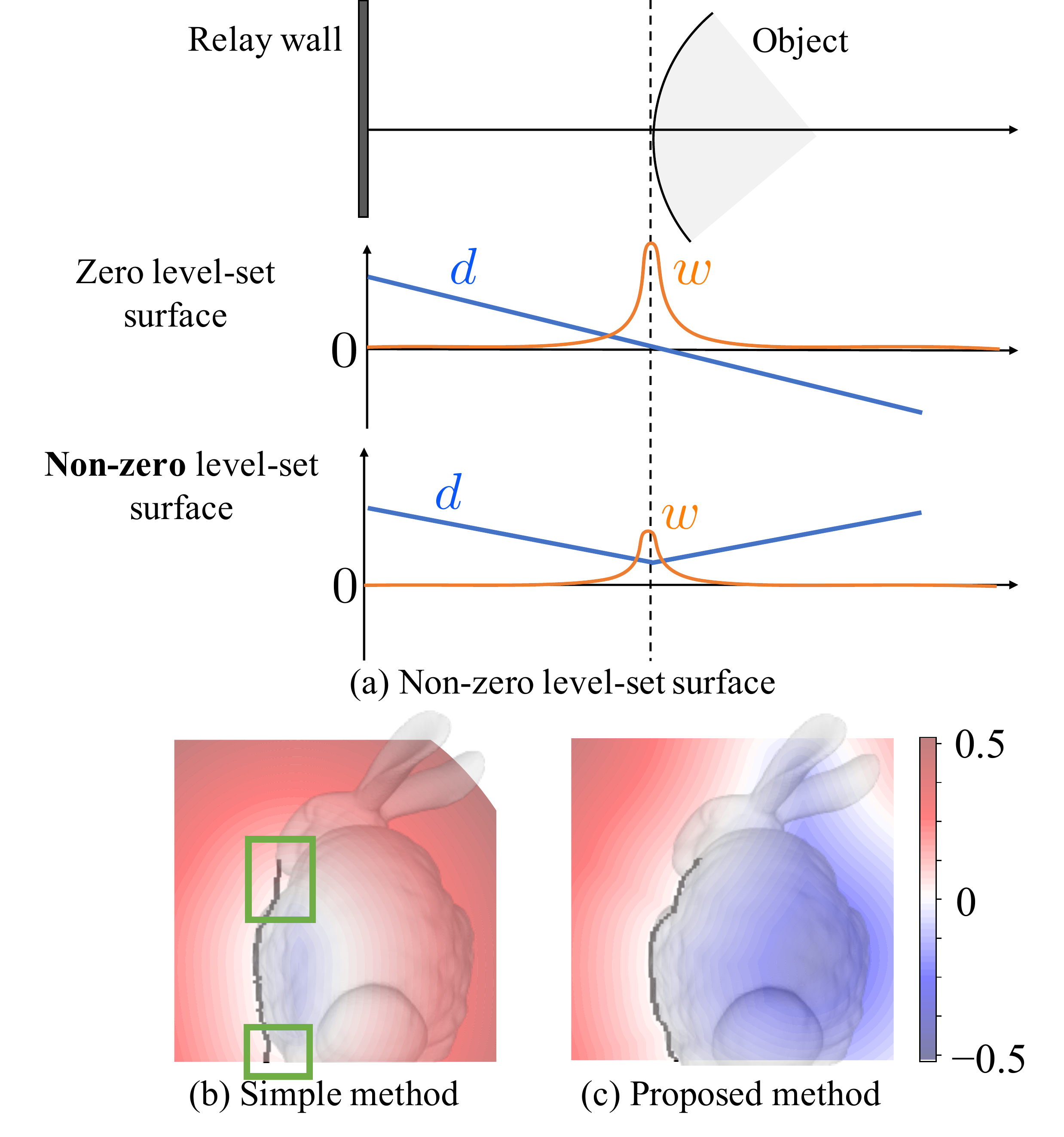}
\caption{(a) Non-zero level-set surface. Under-constrained NLOS setup causes non-zero level-set surface because back side of object cannot be observed from relay wall. Point with non-zero signed distance incorrectly has highest weight on ray from relay wall. 
(b) Learned SDF of synthetic bunny scene, where points with highest weights (black curve) do not coincide with zero level-set in green boxes.
(c) Our method enabled us to extract plausible zero level-set surfaces.
}
\label{fig:nonzero_level_set}
\end{figure}

Many methods for learning neural implicit surfaces have been proposed under a multi-view setup \cite{Niemeyer2020,Oechsle2021,Wang2021neus,Yariv2021,Yariv2020}.
One of the differences between these methods and our NLOS-NeuS is the configuration of the target scene.
In a multi-view setup, the target object is surrounded by multiple cameras.
In an NLOS setup, however, only one side of the target object is observed from the relay wall.
We found that such an under-constrained setup causes a non-zero level-set surface in an SDF, as shown in Fig. \ref{fig:nonzero_level_set}(a).
The back side of the object cannot be observed, which leads to the highest weight at the point with non-zero signed distance on the ray.
Figure \ref{fig:nonzero_level_set}(b) shows the estimated signed distance for the synthetic bunny scene with the simple extension of the NeTF.
The black curve indicates an extracted object surface as the points with the highest weight along the ray from the relay wall and does not coincide with the zero level-set in the green boxes.
Figure \ref{fig:ablation_loss}(a) is the estimated depth from this SDF with the sphere tracing algorithm \cite{Hart1996}, where most parts of the bunny are missing.

To avoid such a non-zero level-set surface, the following two requirements should be satisfied. (1) At the object surface, the signed distance values should be zero. To achieve this, we have to detect the object surface and force explicit geometric supervision. (2) Points with non-zero signed distance should not contribute to the object surface. To achieve this, $\alpha$ in Eq. (\ref{eq:sdf2density}) decreases correctly during training. As explained in Sec. \ref{sec:nlos-surface}, $\alpha \to 0$ means a perfectly sharp surface, i.e., only points with zero signed distance contributes to object surfaces. We introduce two constraints as training losses for satisfying these two requirements, which enabled us to estimate a plausible SDF, as shown in Fig. \ref{fig:nonzero_level_set}(c).

\paragraph{Zero signed-distance self-supervision}
We supervise the MLP to have zero signed distances at the object surface.
Recent studies have shown that such explicit geometric supervision enhances the quality of the learned SDF \cite{Azinovic2022,Fu2022}.
Differing from these studies in which geometric information was obtained with additional sensors, we have no explicit geometric information, thus automatically detect points to be supervised during training.

To detect surface points, we compute the probabilistic density function (PDF) as normalized $w(\mathbf{p},\mathbf{v})\rho(\mathbf{p},\mathbf{v})$ on each scan sphere during training and sample $N_z$ points on the sphere on the basis of the PDF.
These points are likely to be located on the object surface; thus, we force these points to have a zero signed distance with the following loss function:
\begin{equation}
\mathcal{L}_{z} = \frac{1}{MBN_z}\sum_{t,\mathbf{p}'} \sum_{n = 1}^{N_z} m(\mathbf{p}',t) | d(\mathbf{p}(t, \theta_n, \phi_n)) |,
\end{equation}
where $m(\mathbf{p}',t) \in \{ 0,1 \}$ is a mask; 
$m(\mathbf{p}',t)=1$ means that object points exist on the sphere with a radius $r_t$ centered at $\mathbf{p}'$.
This mask can be computed easily by thresholding the measured transient $\tau_m$.
If we model background effects (Sec. \ref{sec:background}), we compute the mask with an object component $\tau_m - \xi\tau_b$ during training.

\paragraph{Constraint on volume-rendering weight}
Wang et al. \cite{Wang2021neus} mentioned that a learnable parameter connecting signed distance and density ($\alpha$ in Eq. (\ref{eq:sdf2density}) in our case) converges to zero during training.
However, we found that an additional constraint is necessary to correctly decrease the parameter for the NLOS setup.

Before introducing the constraint, we discuss a mask loss proposed in multi-view settings \cite{Wang2021neus,Yariv2020}.
In NeuS \cite{Wang2021neus}, the input images are masked, and the masks are used for an additional loss, where the sum of the volume rendering weights (Eq. (\ref{eq:weight}) in our case) on each camera ray is forced to be equal to the value at the corresponding pixel of the mask, i.e., the mask loss is defined as 
\begin{equation}
\mathcal{L}_{mask} = E(m_k, \hat{o}_k), \label{eq:mask_loss}
\end{equation}
where $m_k$ is the value of the mask at the $k$-th pixel, $\hat{o}_k$ is the accumulated weight, and $E$ is an error metric such as the binary cross entropy.
We provide the following observation.
\begin{pro} \label{th:mask_loss}
Minimizing the mask loss (Eq. (\ref{eq:mask_loss})) leads to the convergence of $\alpha$ (in Eq. (\ref{eq:sdf2density})) to 0.
\end{pro}
We provide the proof in the supplementary material.
Intuitively, all densities in the empty space should be 0, which results in the convergence of $\alpha$ to 0.

Therefore, the mask loss is effective for learning $\alpha$, while the difficulty in the NLOS setup is that we cannot obtain an object mask.
Instead of using such explicit mask loss, we use the following loss:
\begin{align}
\mathcal{L}_{en} = \frac{1}{M|\theta||\phi|}\sum_{\mathbf{p}',\theta,\phi}-\hat{o} \log_2 \hat{o} - (1-\hat{o}) \log_2 (1-\hat{o}),
\end{align}
where $\hat{o} = \sum_{t=t_{min}}^{t_{max}}w(\mathbf{p}, \mathbf{v})$ is the accumulated weight for each view line from the relay wall, and $|\cdot|$ is the number of sampled variables.
This loss function is similar to the beta-distribution constraint of accumulated opacity used in the previous studies \cite{Bi2020,Lombardi2019,Mu2022} where accumulated opacity from a camera should be 0 or 1.
We use entropy as this loss function that bounds the loss value between 0 and 1.

\begin{figure}[tb]
\centering
\includegraphics[width=0.4\textwidth]{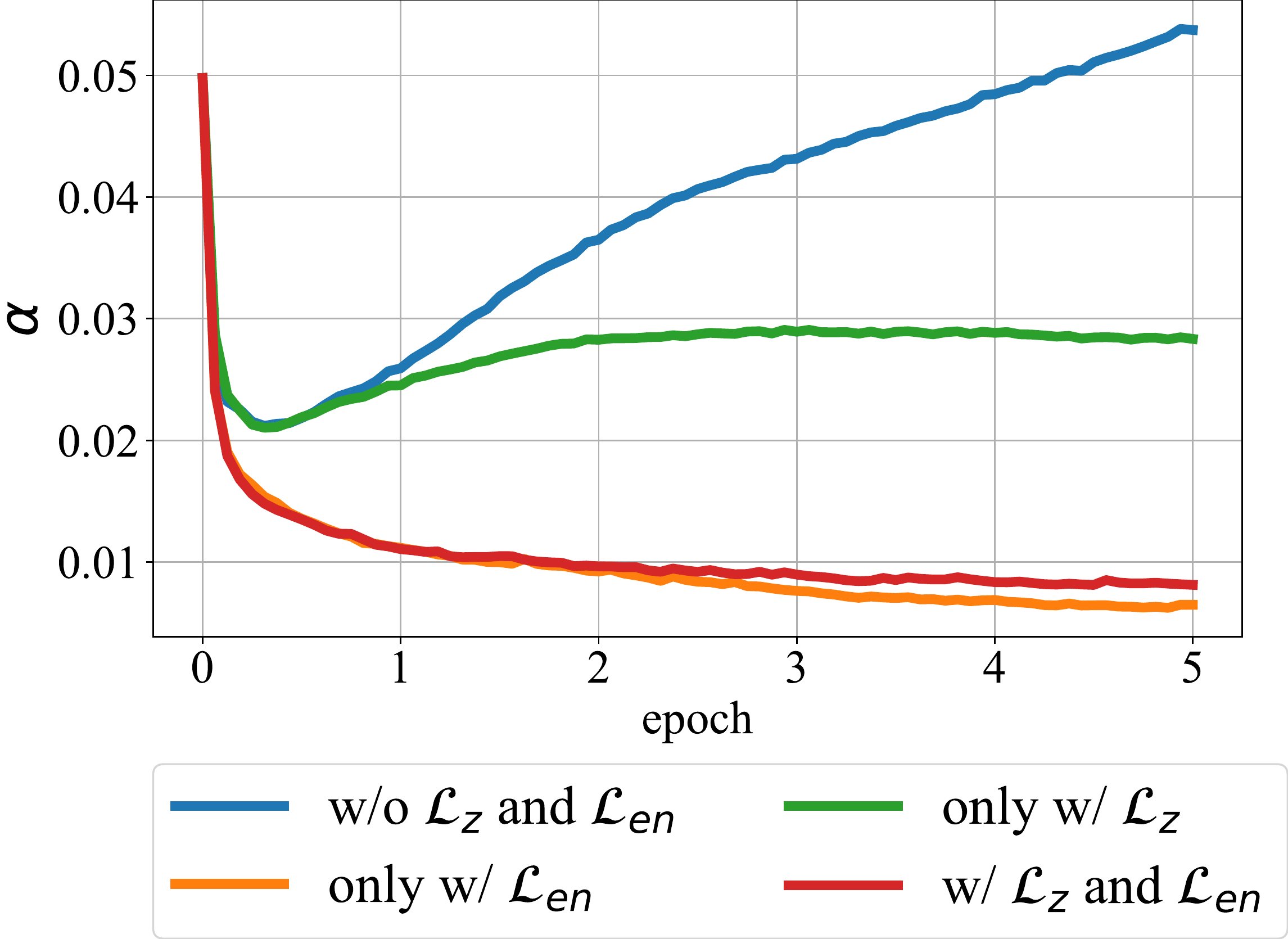}
\caption{Plots of $\alpha$ during training. For learning sharp surface, $\alpha$ in Eq. (\ref{eq:sdf2density}) should decrease during training. This plot shows that $\mathcal{L}_{en}$ can make $\alpha$ converge to very small value (orange and red plots)}
\label{fig:alpha_plot}
\end{figure}

Figure \ref{fig:alpha_plot} shows the plots of $\alpha$ during training for the synthetic bunny scene, where $\alpha$ diverges without both $\mathcal{L}_z$ and $\mathcal{L}_{en}$ (blue plot).
Although $\alpha$ does not diverge if we use $\mathcal{L}_z$ (green plot), it begins to increase in the early stage and remains relatively large at the end of the training.
On the other hand, $\alpha$ successfully decreases and converges to a very small value with $\mathcal{L}_{en}$ (orange and red plots).

\subsection{Training loss}
In summary, we use the following loss function for training NLOS-NeuS:
\begin{equation}
\mathcal{L} = \lambda_\tau \mathcal{L}_\tau + \lambda_{ei} \mathcal{L}_{ei} + \lambda_{z}\mathcal{L}_z + \lambda_{en}\mathcal{L}_{en} + \lambda_f \mathcal{L}_{f},
\end{equation}
where $\mathcal{L}_\tau$, $\mathcal{L}_z$, and $\mathcal{L}_{en}$ are those described in the previous sections.
$\lambda_*$ are hyperparameters to control the contribution of each loss.
$\mathcal{L}_{ei}$ is the Eikonal loss \cite{Gropp2020} for the regularization of the SDF:
\begin{equation}
\mathcal{L}_{ei} = \frac{1}{|\Omega_{\mathit{nlos}}|}\sum_{\mathbf{p}\in \Omega_{\mathit{nlos}}} (\| \nabla d(\mathbf{p}) \| - 1 )^2,
\end{equation}
where $\Omega_{\mathit{nlos}}$ is the target NLOS space.

We also introduce SDF regularization based on the geometry of first-returning photons \cite{Tsai2017}.
From the traveling distances of first-returning photons, we can apply space carving to the NLOS scene to obtain the free space and rough shape.
In the free space, the signed distances to the rough shape form the lower bounds of the true signed distances.
We can leverage these lower bounds as the additional training loss:
\begin{equation}
\mathcal{L}_{f} = \frac{1}{|\Omega_{\mathit{free}}|}\sum_{\mathbf{p} \in \Omega_{\mathit{free}}} \max (0, b(\mathbf{p}) -  d(\mathbf{p})),
\end{equation}
where $\Omega_{\mathit{free}}$
is the free space and $b(\mathbf{p})$ is the lower bound of the signed distance at $\mathbf{p}$ in the free space.
This loss is inspired by the SDF lower bound from silhouette \cite{Lin2020}, while we derive the lower bounds from the geometry of first-returning photons.
In the supplementary material, we provide the details of the detection of the first-returning photons and robust space carving algorithm.

\subsection{Rendering background} \label{sec:background}
In practical scenes, background effects, such as intereflection or reflected light from the floor, are not negligible.
Plack et al. \cite{Plack2023} proposed the background network, which takes as input a point on a relay wall and temporal bin for estimating components from the background.
The scaling parameter between the components from the target object and background is also optimized with an additional constraint.
To avoid an SDF fitting the background components, we follow the same approach for modeling the background but simply adjust the scaling parameter with the measured transient.
Formally, the final transient $\tau$ is the weighted sum of the transients from the object $\tau_{o}$ (same with Eq. (\ref{eq:discrete})) and from the background $\tau_b$ as follows:
\begin{equation}
\tau(\mathbf{p}') = \tau_o(\mathbf{p}') + \xi \tau_b(\mathbf{p}'),
\end{equation}
where $\xi$ is a scaling parameter computed as 
\begin{equation}
\xi = \frac{\sum_t \tau_m(\mathbf{p}',t) - \sum_t \tau_o(\mathbf{p}',t)}{\sum_t \tau_b(\mathbf{p}',t)}.
\end{equation}
The background network is jointly trained with $\rho$ and $d$.
In the supplementary material, we provide examples of the rendered background.
\section{Experiments}
We evaluated the effectiveness of NLOS-NeuS on publicly available NLOS datasets.
We compared it with the NeTF \cite{Shen2021} and directional LCT (DLCT) \cite{Young2020}, which is a state-of-the-art surface-reconstruction method for NLOS scenes.
Note that we retrained the NeTF with the author's code.

\begin{figure*}[tb]
\centering
\includegraphics[width=1.\textwidth]{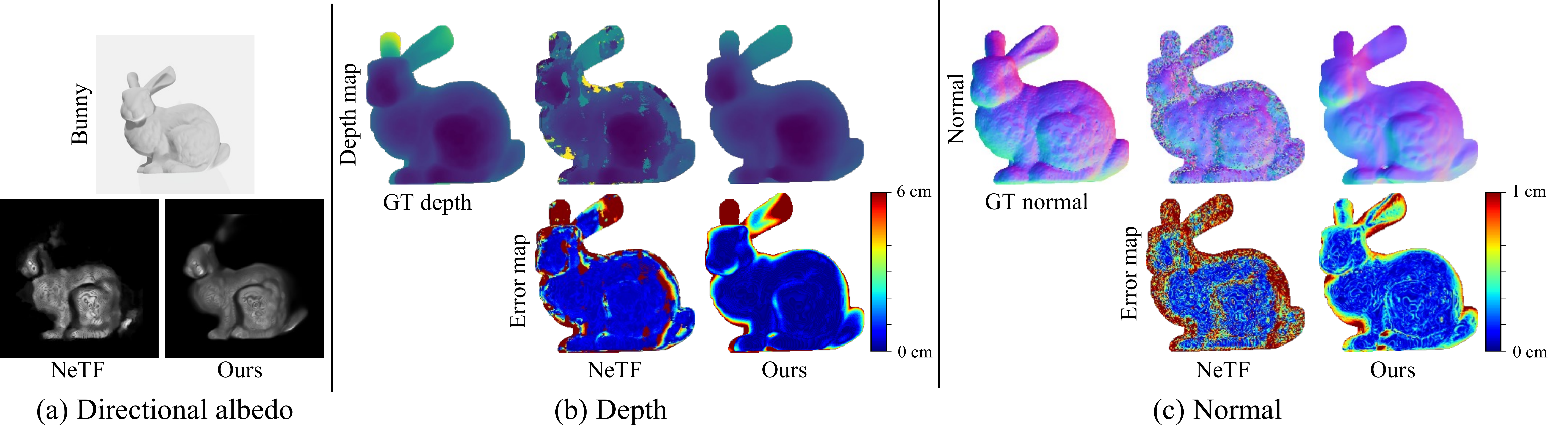}
\caption{Results from NeTF \cite{Shen2021} and NLOS-NeuS on synthetic bunny scene. (a) Rendered directional albedo, (b) depth reconstruction, and (c) surface-normals reconstruction. Neural implicit surface representation achieved geometrically consistent reconstruction.
}
\label{fig:bunny_result}
\end{figure*}

\begin{figure}[tb]
\centering
\includegraphics[width=0.48\textwidth]{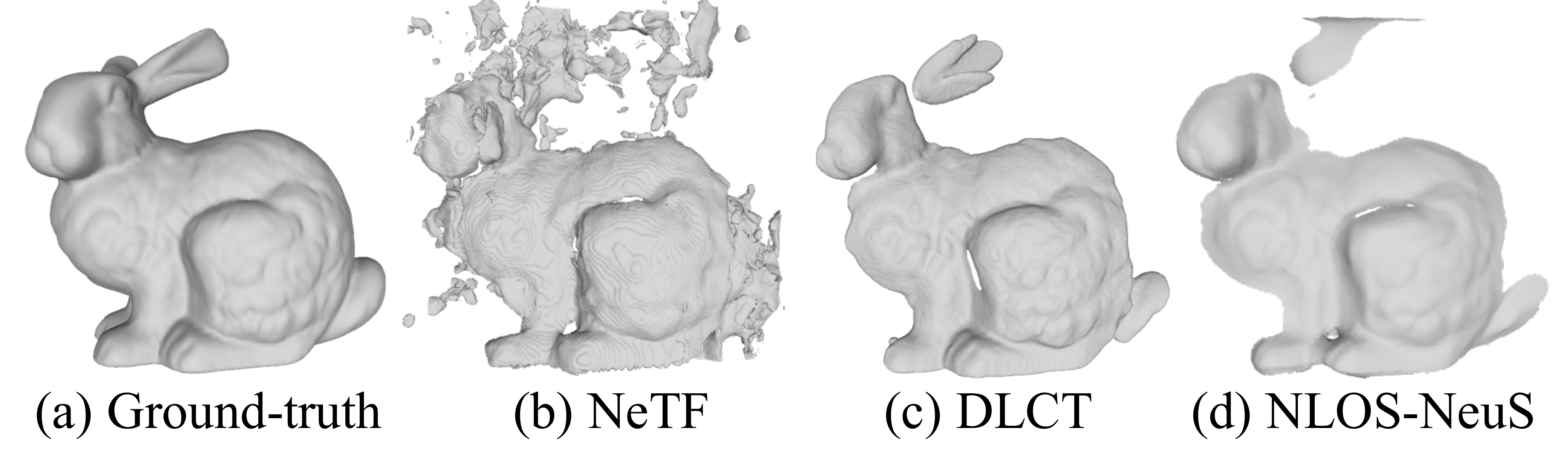}
\caption{Surface reconstruction of bunny. (a) Ground-truth mesh, (b) mesh from NeTF \cite{Shen2021}, (c) mesh from DLCT \cite{Young2020}, and (d) mesh from NLOS-NeuS.
}
\label{fig:bunny_mesh}
\end{figure}

\begin{figure*}[tb]
\centering
\includegraphics[width=0.9\textwidth]{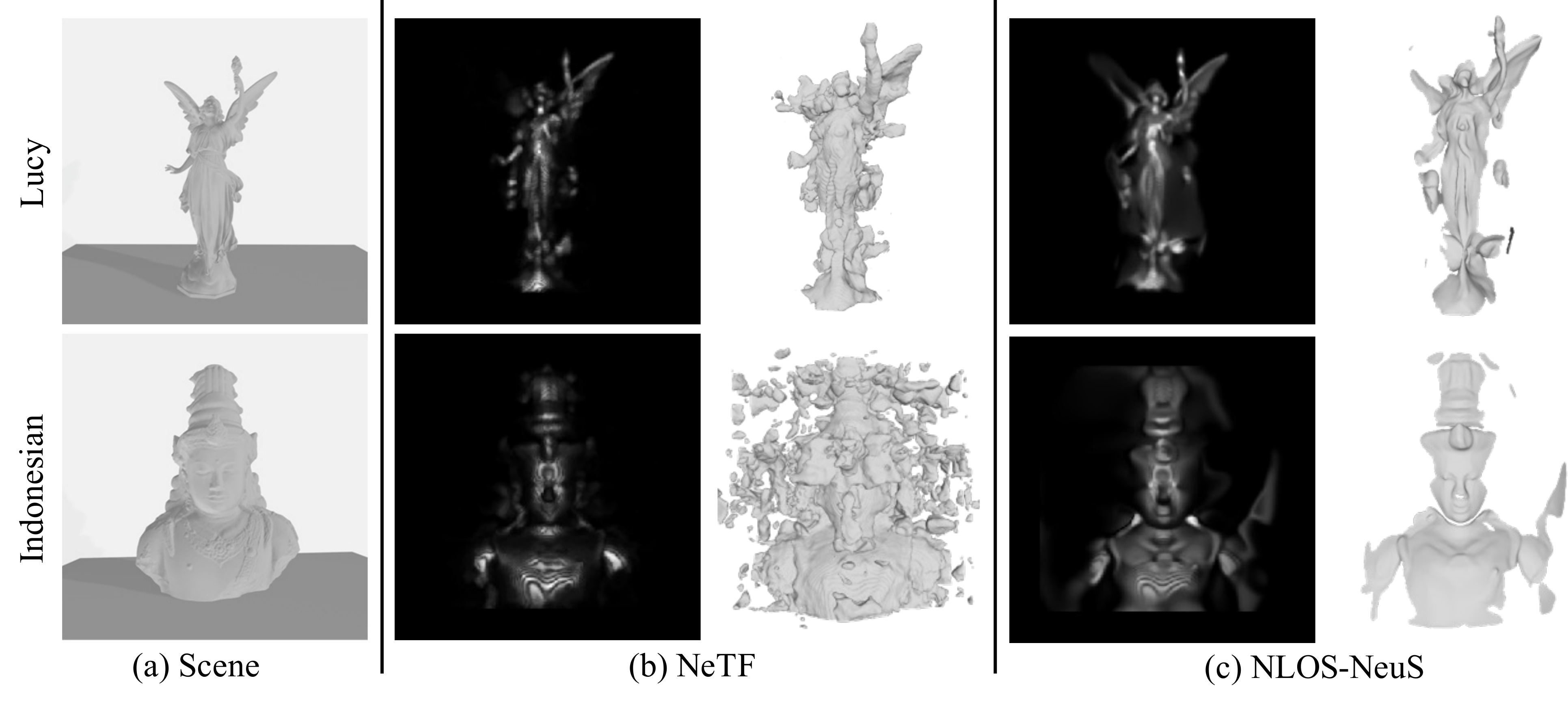}
\caption{Results on Lucy and Indonesian with  (b) NeTF \cite{Shen2021} and (c) NLOS-NeuS. Results contain directional albedos and reconstructed surface. Note that each object is located on floor and measured transient is highly affected by reflected light from floor. }
\label{fig:znlos_result}
\end{figure*}

\begin{figure*}[tb]
\centering
\includegraphics[width=0.95\textwidth]{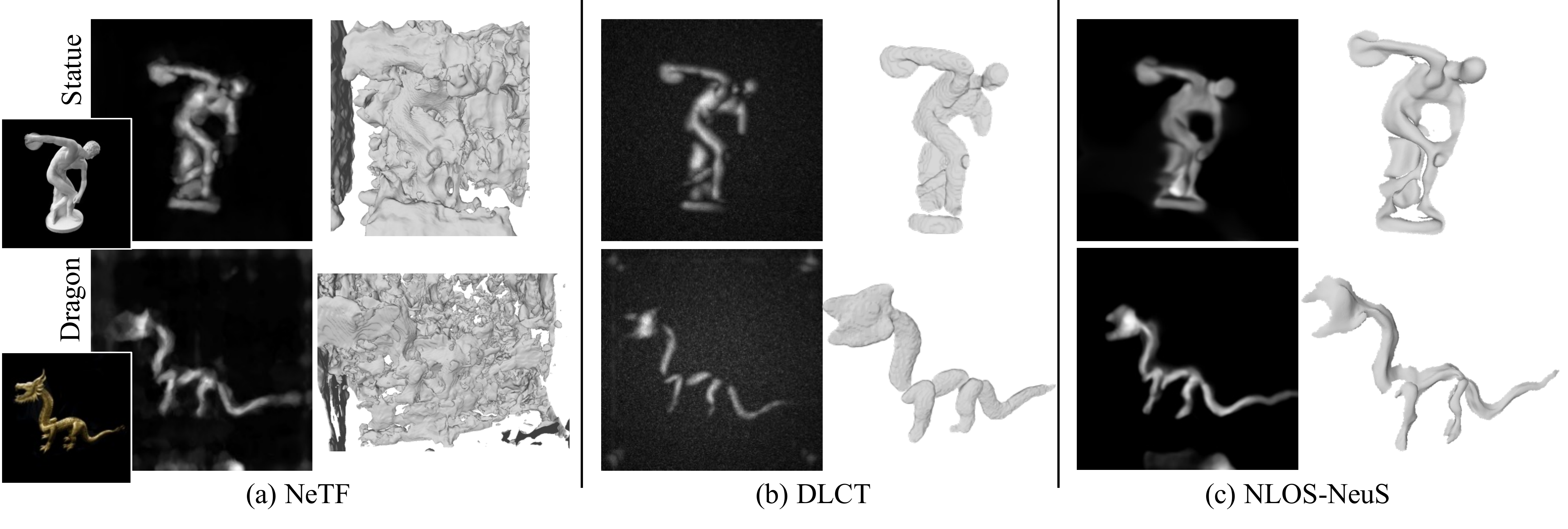}
\caption{Results of surface reconstruction on real dataset with (a) NeTF \cite{Shen2021}, (b) DLCT \cite{Young2020}, and (c) NLOS-NeuS.
}
\label{fig:fk_mesh}
\end{figure*}

\begin{figure*}[tb]
\centering
\includegraphics[width=0.9\textwidth]{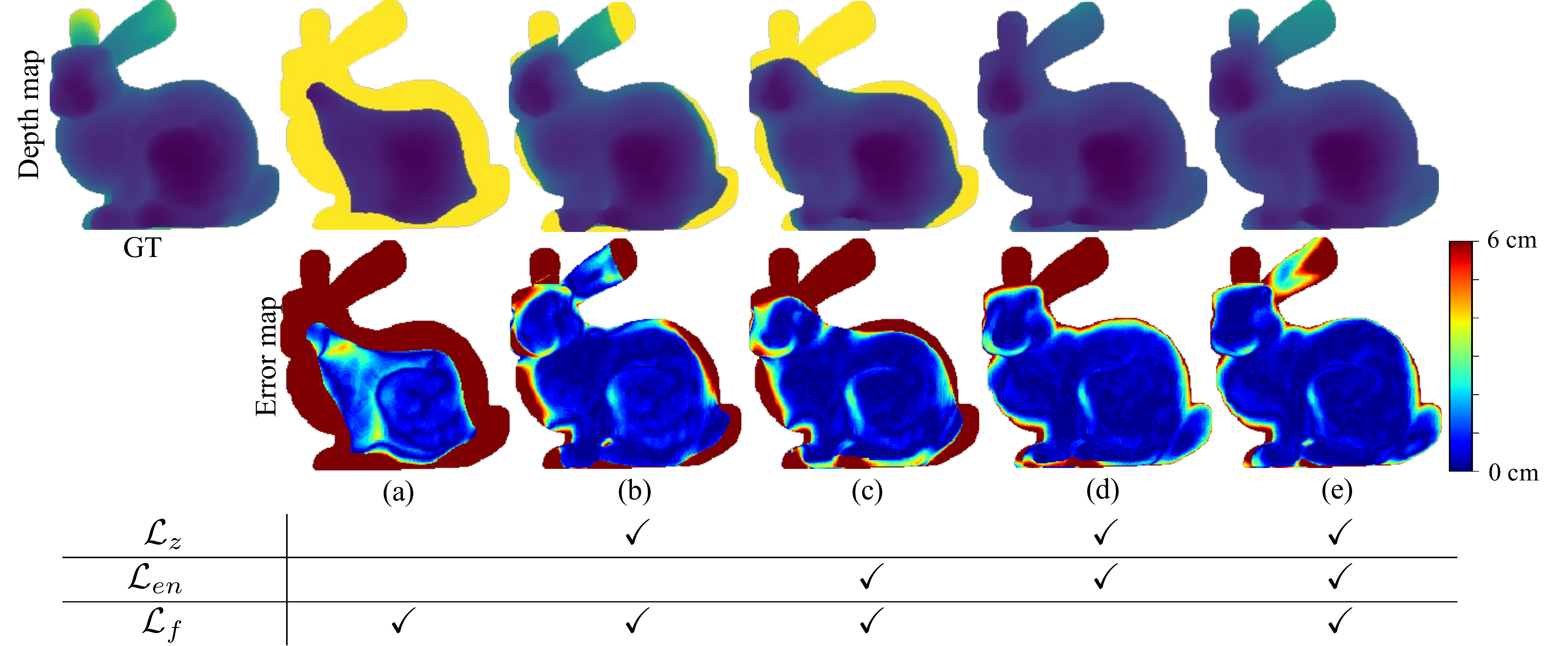}
\caption{Ablation study of loss functions for learning zero level-set surface. By using both  $\mathcal{L}_z$ and $\mathcal{L}_{en}$, MLP can correctly learn zero level-set surface. We also show effectiveness of $\mathcal{L}_f$, which improves overall accuracy of reconstructed geometry.}
\label{fig:ablation_loss}
\end{figure*}

\begin{figure*}[tb]
\centering
\includegraphics[width=0.9\textwidth]{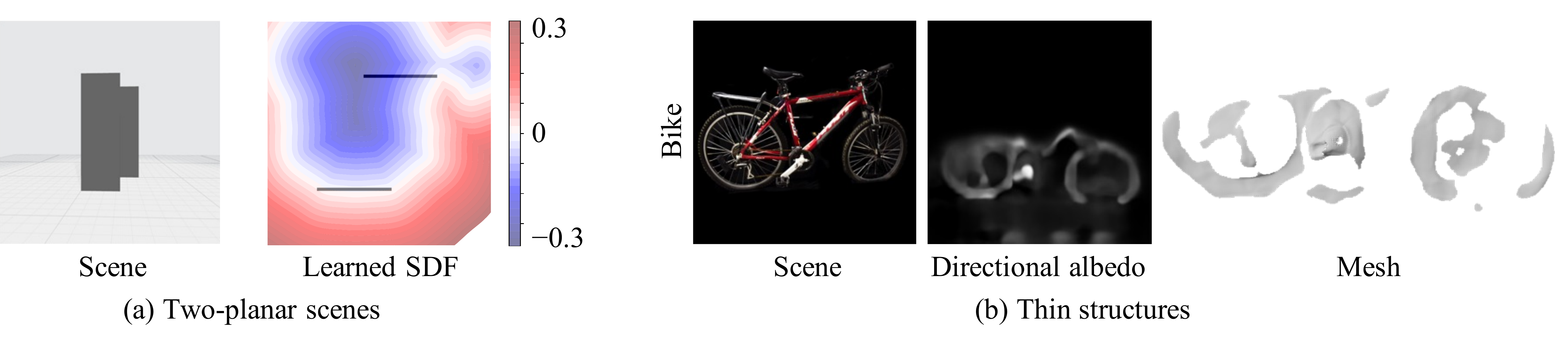}
\caption{Challenging scenes for SDF. (a) This scene consists of two planes, one of which is partially occluded. SDF is not suitable for such non-closed surface. (b) This scene consists of thin structures, which are also difficult for SDF to correctly represent geometry.}
\label{fig:limitation}
\end{figure*}

\subsection{Implementation}

\paragraph{Network architecture} 
We use the same networks with that of NeuS \cite{Wang2021neus} with geometric initialization \cite{Atzmon2020} for $d$ and $\rho$.
The inputs of the networks are a scene position and direction to the relay wall with positional encoding \cite{Mildenhall2020}.
Differing from NeuS, we do not use a surface normal for the input of $\rho$ due to heavy computational cost.
For the background network, we use a three-layers MLP with positional encoding for both position and time.

\paragraph{Surface reconstruction}
After training NLOS-NeuS, we use the sphere tracing algorithm \cite{Hart1996} to extract a point cloud instead of using the marching cubes algorithm \cite{Lorensen1987} because an optimized SDF is not correctly closed due to the NLOS setup.
Note that we first create an object mask with estimated directional albedo then apply the sphere tracing algorithm to only the object region.
After extracting the point cloud with this algorithm, we can obtain a surface normal at each point by evaluating the gradient of $d$.
We then use the Poisson surface reconstruction \cite{Kazhdan2013} to obtain the object surface.

\paragraph{Dataset}
We use two datasets preprocessed in the NeTF \cite{Shen2021}.
The first one is the ZNLOS dataset \cite{Galindo19}, which was synthetically created with the transient renderer \cite{Jarabo2014}.
The other one is the f-k dataset \cite{Lindell2019}, which is a real capture using gated SPADs avoiding strong direct reflection from a relay wall.
The temporal resolution of the SPAD system is approximately 70 pico seconds.

\begin{table}[tb]
\centering
\caption{Quantitative comparison of NLOS-NeuS with NeTF \cite{Shen2021} and DLCT \cite{Young2020} on bunny scene. We computed end-point-error for surface normals. Errors dramatically reduced with NLOS-NeuS compared with NeTF, and NLOS-NeuS is also comparable with DLCT.}
\label{tab:quantitative_comparison}
  \begin{tabular}{ccccc} \toprule
    & \multicolumn{2}{c}{Depth [cm]} & \multicolumn{2}{c}{Normal [cm]} \\
    & RMSE & MAE & RMSE & MAE \\ \midrule
    NeTF \cite{Shen2021} & 8.26 & 3.67 & 0.81 & 0.63 \\
    DLCT \cite{Young2020} & 5.27 & {\bf 1.59} & 0.40 & {\bf 0.30} \\
    NLOS-NeuS & {\bf 4.63} & 1.84 & {\bf 0.39} & {\bf 0.30} \\ \bottomrule
  \end{tabular}
\end{table}

\subsection{Results on synthetic data}
Figure \ref{fig:bunny_result} shows the comparison of the NeTF and NLOS-NeuS on the synthetic bunny scene.
Figure \ref{fig:bunny_result}(a) visualizes the rendered directional albedos with the NeTF and NLOS-NeuS.
The directional albedo of the NeTF was simply extracted as the highest $\sigma \rho$ on a ray from the relay wall.
The directional albedo of NLOS-NeuS was generated with volume rendering.
In the supplementary material, we
present our directional albedos rendered from different views, which is similar to novel-view synthesis with the NeRF \cite{Mildenhall2020}.
Figure \ref{fig:bunny_result}(b) shows the results of depth reconstruction.
Note that the results are masked with the ground-truth depth map.
The neural implicit surface representation with NLOS-NeuS dramatically improved the accuracy of the depth reconstruction.
We also show the reconstructed surface normals in (c), where the NeTF normal is computed as the normalized gradient of the density network, demonstrating the geometrically consistent reconstruction of NLOS-NeuS.
Table \ref{tab:quantitative_comparison} shows the quantitative comparison with the NeTF and DLCT \cite{Young2020} on depth and surface normals.
The errors dramatically reduced with NLOS-NeuS compared with the NeTF, and NLOS-NeuS is also comparable with the DLCT, which is the state-of-the-art surface reconstruction in NLOS scenes.

Figure \ref{fig:bunny_mesh} shows the qualitative comparison of reconstructed meshes with the NeTF, DLCT and NLOS-NeuS.
Due to the lack of geometric representation, the NeTF mesh contains large errors.
On the other hand, NLOS-NeuS enables high quality 3D surface reconstruction.
Although the result with the DLCT is comparable with that with NLOS-NeuS on this synthetic data, its discretized representation is limited in real data, as shown in Fig. \ref{fig:nlos_setup}(c).

Figure \ref{fig:znlos_result} summarize the comparison with the NeTF for the Lucy and Indonesian scenes.
Note that each object is located on a floor and the measured transient is highly affected by the reflected light from the floor.
In the supplementary material, we present the learned transients of the floor reflection.
We present the directional albedos and meshes of both methods.
NLOS-NeuS can reconstruct the complicated structure in the Lucy.
Although there are some missing parts in our results of the Indonesian, NLOS-NeuS can obtain geometrically plausible results compared with the NeTF.

\subsection{Results on real capture}
Figure \ref{fig:fk_mesh} shows the results with the real data captured using a SPAD.
We present the directional albedos and reconstructed meshes for both scenes.
The NeTF fails to reconstruct the correct geometry in these scenes.
DLCT can reconstruct the object structures, while the quality of the reconstructed geometry is limited due to its discrete representation.
NLOS-NeuS, on the other hand, can reconstruct the fine details and smooth surface compared with DLCT because of its continuous representation of the neural implicit surface.
These results indicate the effectiveness of NLOS-NeuS for real scenes.

\subsection{Ablation study on loss functions}
We introduce several losses for training an SDF.
Figure \ref{fig:ablation_loss} shows the comparison between the several patterns of the loss functions on the bunny.
As discussed in Sec. \ref{sec:nonzero_levelset}, the simple extension of the NeTF with the neural implicit surface incorrectly estimates non-zero level-set surfaces.
When we train the MLP without both $\mathcal{L}_z$ and $\mathcal{L}_{en}$ (a), most parts of the shape are missing due to such non-zero level-set surfaces.
When we train the MLP only with $\mathcal{L}_z$ (b) or $\mathcal{L}_{en}$ (c), the results dramatically improve, while the shape near the object boundary cannot be reconstructed.
By using both loss functions (e), we can correctly estimate the shape near the object boundary.
We also show the effectiveness of $\mathcal{L}_f$, which improves the overall accuracy of the reconstructed geometry (d,e).

\subsection{Limitations}
We finally discuss the limitations due to our geometric representation.
Figure \ref{fig:limitation} shows an example of a challenging scene.
In Fig. \ref{fig:limitation}(a), two planes are located and one is partially occluded. 
The figure also shows learned SDF, where the black lines indicate the positions of the ground-truth planes.
Although the zero level-set is extracted on the front plane, the SDF around the back plane is not correct.
Such a non-closed surface is difficult for an SDF to represent the geometry.
Figure \ref{fig:limitation}(b) is another example of a challenging scene for an SDF, where the target object consists of thin structures.
One approach for these scenes is to use more flexible geometric representation such as unsigned distance field \cite{Chibane2020udf}.

\section{Conclusion}
We proposed NLOS-NeuS, which is an extension of the NeTF for representing 3D surfaces with an SDF in NLOS scenes. 
To the best of our knowledge, this is the first study on neural implicit surfaces with volume rendering in NLOS scenes.
The experimental results indicate that the introduced constraints are essential for learning a correct SDF to avoid non-zero level-set surfaces in an under-constrained NLOS setup, and NLOS-NeuS with continuous representation enables high quality 3D surface reconstruction in NLOS scenes.
Future research directions include using other geometric representations and an extension for a non-confocal NLOS setup.

{\small
\section*{Acknowledgement}
This work was supported by the Japan Society for the Promotion of Science KAKENHI Grant Number 21K21317, 20K20629.
}

{\small
\bibliographystyle{ieee_fullname}
\bibliography{egbib}
}

\newpage

\section*{Supplementary material}
\appendix

\begin{figure}[tb]
\centering
\includegraphics[width=0.4\textwidth]{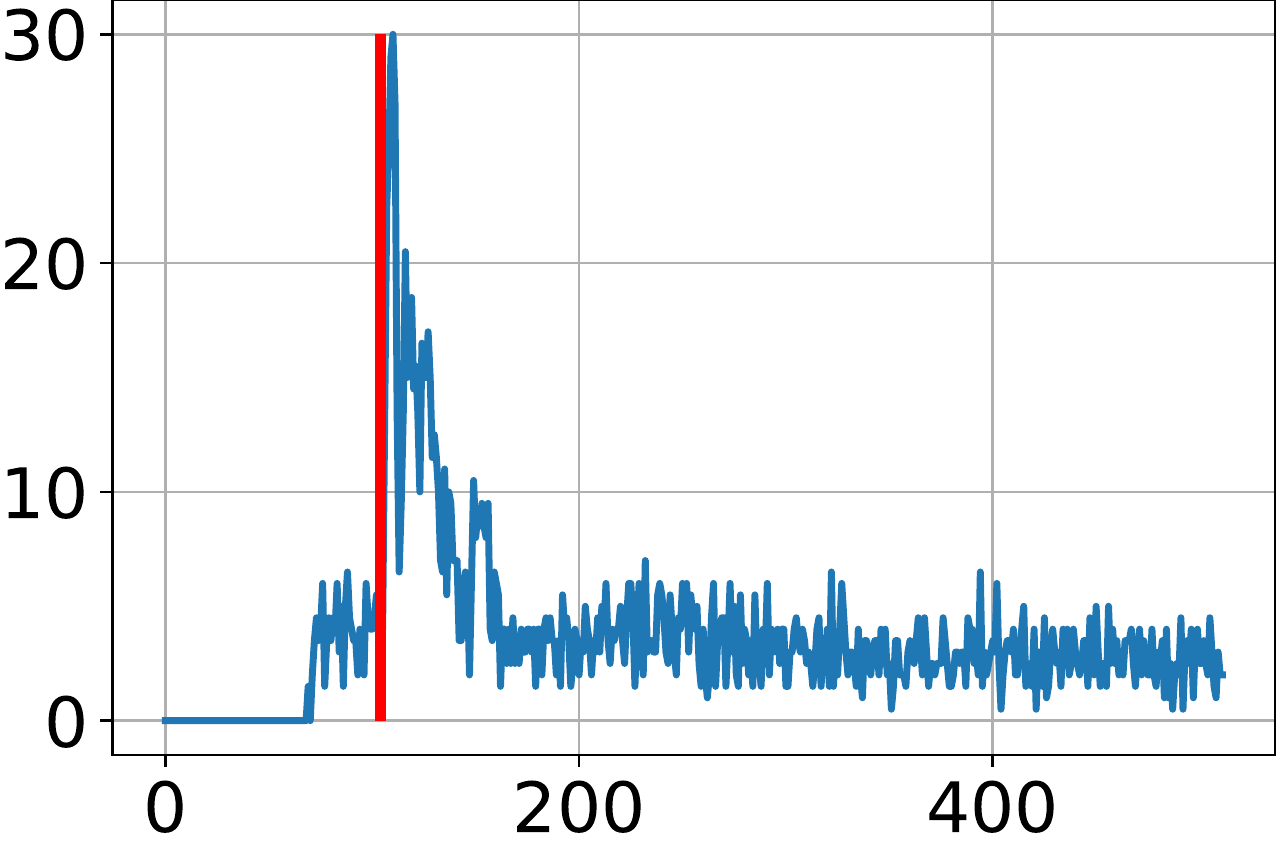}
\caption{Example of detecting temporal bin corresponding to first-returning photon in real data. Blue plot is measured transient and red line indicates detected temporal bin.}
\label{fig:first_returning_photon}
\end{figure}

\begin{figure*}[tb]
\centering
\includegraphics[width=0.9\textwidth]{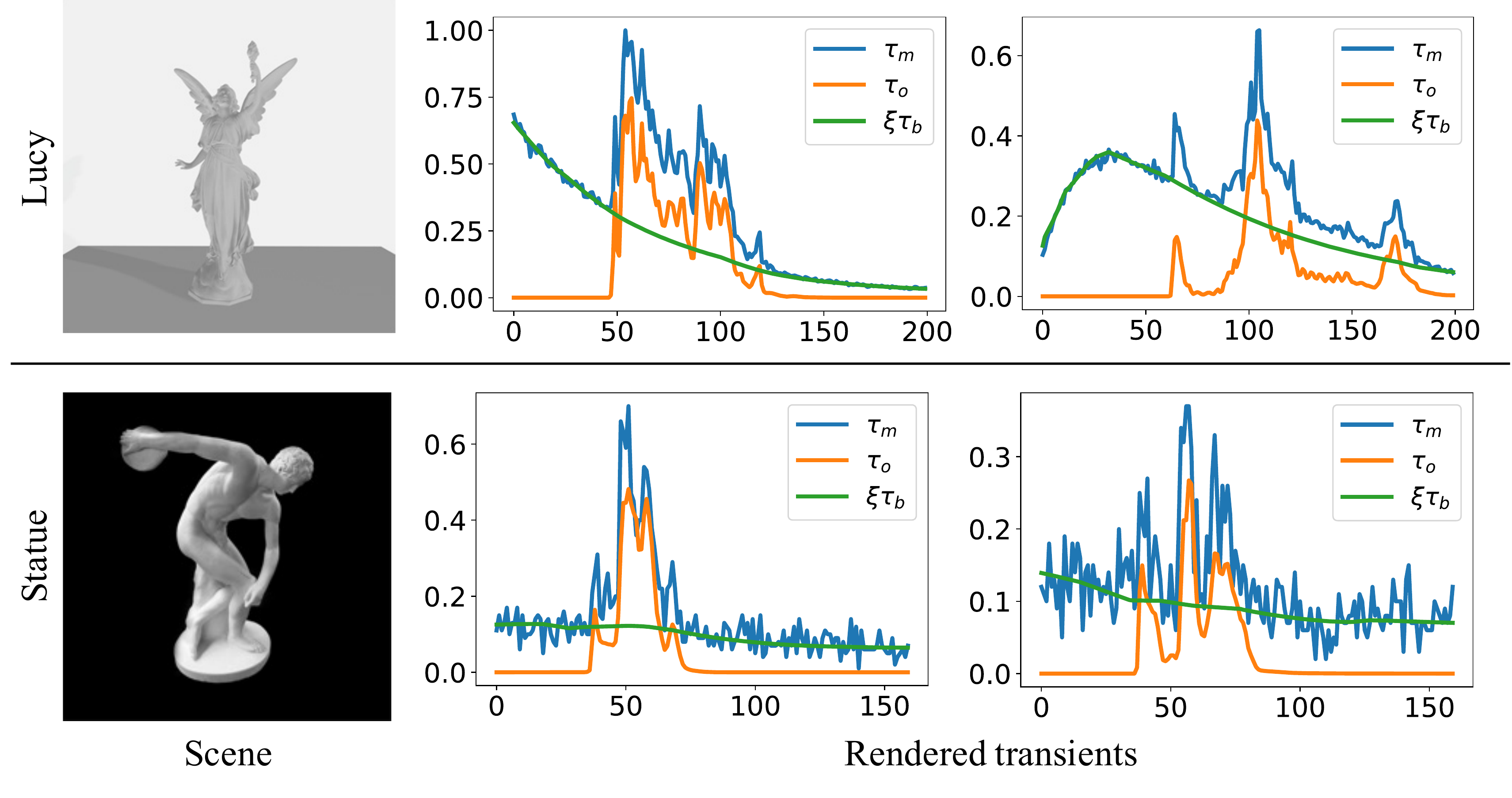}
\caption{Examples of background rendering. Blue, orange, and green plots indicate measured transients and components from object and background, respectively.}
\label{fig:background_hist}
\end{figure*}

\section{Proof of Proposition \ref{th:mask_loss}}
This section provides the proof of the Proposition \ref{th:mask_loss}.
We rewrite Eq. (\ref{eq:sdf2density}) as
\begin{align}
f(x) = \frac{1}{\alpha} \frac{1}{1+e^{\frac{x}{\alpha}}}.
\end{align}

\begin{proof}
To analyze the behavior of the mask loss (Eq. (\ref{eq:mask_loss})), we consider a sampling ray near the object boundary.
In the empty space, all densities at sampled points on the ray should be zero.
Now let $\epsilon > 0$ be the minimum signed distance value on this ray, which also leads to the highest density on the ray and should be close to zero in the empty space, i.e., $f(\epsilon)\to 0$.

In the empty space, the signed distance at each position is positive.
When $x>0$, the first and second derivatives of $f(x)$ satisfy $f'(x) < 0$ and $f''(x) > 0$.
Thus, $f(\epsilon)$ can be lower bounded by the first-order derivative approximation around $x=0$:
\begin{align}
f(\epsilon) &> f(0) + f'(0)\epsilon = \frac{1}{2\alpha} - \frac{1}{4 \alpha^2}\epsilon.
\end{align}
Thus, $f(\epsilon)\to 0$ leads to $\alpha < \epsilon/2$.
When minimizing the mask loss, $\alpha$ can be upper bounded by $\epsilon/2$.
If points are sampled densely enough, at the object boundary where $\epsilon \to 0$, $\alpha$ converges to 0.
\end{proof}

\section{Space carving with the first-returning photons}
For the free space loss $\mathcal{L}_{f}$, we apply the space carving algorithm on the basis of the geometry of first-returning photons \cite{Tsai2017}.
In this section, we explain the detection of the temporal bin corresponding to the first-returning photon and robust space carving algorithm.

\subsection{Detection of the first-returning photons}
For the detecting the first-returning photons, we simply compute the difference between the values of nearby temporal bins in a transient.
Formally, let $\tau_m \in \mathbb{R}^B$ be a measured transient.
At each temporal bin $t$, we compute $\tau_m(t) - \tau_m(t+1)$.
The bin of the first-returning photon is then computed as 
$\arg \min_{t \in S} [ \tau_m(t) - \tau_m(t+1) ],\; S =\{t | \tau_m(t) - \tau_m(t+1) > \eta \}$, where $\eta$ is a threshold value.
However, real data captured using a SPAD contains high-frequency noise and background effects; thus, we apply the Gaussian filter and thresholding to suppress such undesirable signals as preprocessing.
Figure \ref{fig:first_returning_photon} shows an example of the detected bin of the first-retuning photon in the real data.

\subsection{Robust space carving}
After detecting the temporal bins of the first-returning photons, we can apply the space carving algorithm \cite{Tsai2017}.
Let a detected temporal bin be $t$ and the corresponding radius be $r_t$ at a relay wall position $\mathbf{p}'$.
It is guaranteed that an object does not exist in the sphere centered at $\mathbf{p}'$ with $r_t$.
We propose a robust space carving method since the detected temporal bins have some errors due to undesirable signals such as noise.
First, the target space is divided into $128 \times 128 \times 128$ grid voxels $\mathcal{V}$.
We vote for a voxel $v \in \mathcal{V}$ if $v$ is inside each carving sphere.
Let $c(v)$ be the total number of votes of a voxel $v$ after voting with all spheres.
Instead of simply using the carved space, we compute an object space $\Omega_{obj}$ and free space $\Omega_{free}$ as $\Omega_{obj} = \{v | c(v) > 0.99 \times \max_{v' \in \mathcal{V}} c(v') \}$ and $\Omega_{free} = \mathcal{V} - \Omega_{obj}$ for robust space carving.

\section{Training details}\label{sec:implementation}
We set the weights of the training loss as $[\lambda_\tau, \lambda_{ei}, \lambda_z, \lambda_{en}, \lambda_f] = [1., 0.1, 0.01, 0.001, 0.01]$ except for the scene ``bike,'' where we set $\lambda_{en} = 0.002$ because we found that the scene consisting of thin structures requires large $\lambda_{en}$ for reducing $\alpha$ during training.

When rendering a transient, we sampled 64 angles for both $\theta$ and $\phi$ on each scan sphere.
For $\mathcal{L}_{ei}$, we randomly sampled 4096 points in a target NLOS space at each iteration.
For $\mathcal{L}_f$, we randomly sampled 4096 voxels in a free space at each iteration.

The optimizer was Adam \cite{Kingma2015} with the hyperparameters $lr=1.0 \times 10^{-4}$, $\beta_1=0.9$, and $\beta_2=0.999$.

\section{Examples of background rendering}
To model background effects in measured transients, a shallow MLP for rendering the background effects is jointly trained with an SDF as explained in Sec. \ref{sec:background}.
Figure \ref{fig:background_hist} shows examples of rendering background effects.
In the synthetic Lucy scene, the object is located on a floor, and the measured transient contains reflection from the floor.
The background network correctly divides the transient into object and floor components.
In the real statue scene, the measured transient has the background effects with high-frequency noise.
Although the characteristics of these background effects are different, 
the background network correctly fit both effects.

\section{Rendering from different views}
\begin{figure*}[tb]
\centering
\includegraphics[width=0.9\textwidth]{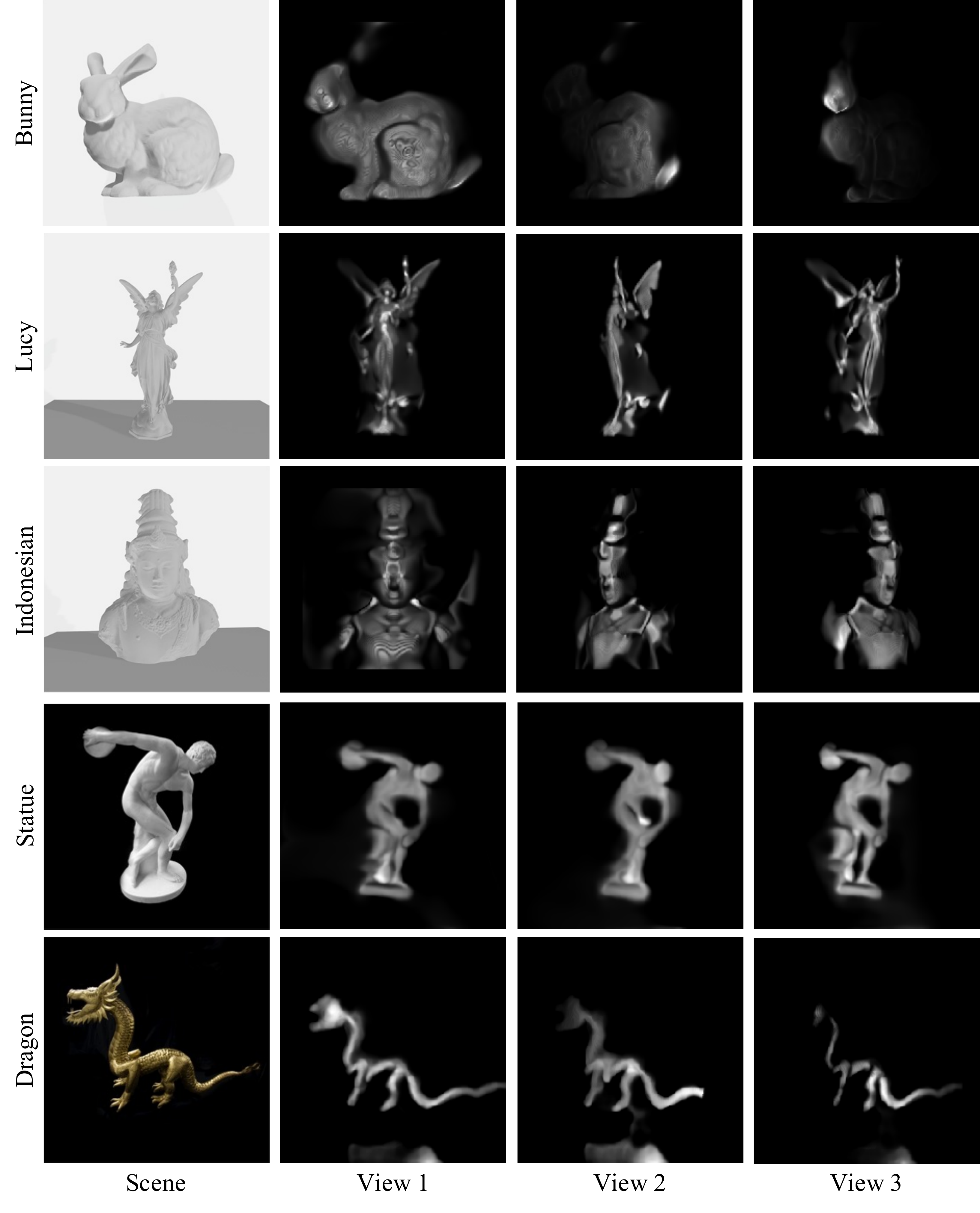}
\caption{Rendered directional albedos from 3 different views}
\label{fig:different_views}
\end{figure*}

Our neural implicit representation can render directional albedos from different views with volume rendering after training, similar to novel-view synthesis with the NeRF \cite{Mildenhall2020}.
Figure \ref{fig:different_views} shows examples of rendered directional albedos from three different views.

\end{document}